\pdfoutput=1

\documentclass[11pt]{article}

\usepackage{ACL2023}

\usepackage{times}
\usepackage{latexsym}
\usepackage{booktabs}
\usepackage{graphicx}
\usepackage[T1]{fontenc}

\usepackage[utf8]{inputenc}

\usepackage{microtype}

\usepackage{inconsolata}

\title{Duluth at SemEval-2025 Task 7: TF-IDF with Optimized Vector Dimensions for Multilingual Fact-Checked Claim Retrieval}

\author{
    \textbf{Shujauddin Syed} \& \textbf{Ted Pedersen} \\
    Department of Computer Science \\
    University of Minnesota \\
    Duluth, MN 55812 USA \\
    \texttt{\{syed0093, tpederse\}@d.umn.edu}
}

\begin{document}
\maketitle
\begin{abstract}
This paper presents the Duluth approach to the SemEval-2025 Task 7 on Multilingual and Crosslingual Fact-Checked Claim Retrieval. We implemented a TF-IDF-based retrieval system with experimentation on vector dimensions and tokenization strategies. Our best-performing configuration used word-level tokenization with a vocabulary size of 15,000 features, achieving an average success@10 score of 0.78 on the development set and 0.69 on the test set across ten languages. Our system showed stronger performance on higher-resource languages but still lagged significantly behind the top-ranked system, which achieved 0.96 average success@10. Our findings suggest that though advanced neural architectures are increasingly dominant in multilingual retrieval tasks, properly optimized traditional methods like TF-IDF remain competitive baselines, especially in limited compute resource scenarios.

\end{abstract}

\section{Introduction}

The SemEval-2025 Task 7 on Multilingual and Crosslingual Fact-Checked Claim Retrieval addresses the challenge of identifying previously fact-checked claims across multiple languages \cite{semeval2025task7}. This task is important because the global spread of misinformation makes it difficult for professional fact-checkers to manually identify existing fact-checks, as false narratives frequently cross linguistic boundaries. The task covers a diverse set of languages including English, Spanish, German, Portuguese, French, Arabic, Malay, Thai, and two surprise languages (Polish and Turkish) in the test phase, creating a broad evaluation framework for multilingual information retrieval systems in the fact-checking domain.

Our approach uses a TF-IDF\footnote{ \url{https://scikit-learn.org/stable/modules/generated/sklearn.feature_extraction.text.TfidfVectorizer.html}.}  \cite{sparck1972statistical} based retrieval system with optimization of key parameters. We experimented with different vector dimension settings and tokenization to identify an optimal configuration for multilingual retrieval. After testing, we found that a word-level tokenization approach with a vocabulary size of 15,000 features provided the best performance across all languages. Our pipeline\footnote{Our code is publicly available at \url{https://github.com/syed0093-umn/SemEval_Task7}.} includes data cleaning, text preprocessing, vector embedding generation, similarity computation, and ranked retrieval of the most relevant fact-checked claims for each social media post.

The optimized TF-IDF system achieved an average success@10 score of 0.78 on the development set and 0.69 on the test set, ranking 23rd out of 28 participating teams. The top-performing system outperformed our approach with an average score of 0.96, showing the performance gap between traditional statistical methods and modern neural approaches.  Our system performed better on languages such as French: 0.814 and Arabic: 0.820, but struggled with others such as English: 0.452 and Spanish: 0.546. These results suggest that though TF-IDF can serve as a reasonable baseline, significant improvements in multilingual retrieval tasks require advanced contextual embedding techniques and language-specific optimizations.


\section{Task Description}

The SemEval-2025 Task 7 organizers offer the \textbf{MultiClaim} dataset, an augmented and modified version of the original MultiClaim corpus~\cite{Pikuliak_2023}. This dataset collects social media posts containing potential misinformation, previously fact-checked claims, and various related information. The following information in the dataset is used in this task:

\begin{itemize}
    \item \texttt{post/query}: The social media post containing a potential misinformation claim, for example : 
\textit{"[(1525826671.0, 'fb')]","[(""fb/david avocado wolfe Flip the bell peppers over to check their gender. The ones with four bumps are female and those with three bumps are male. The female peppers are full of seeds, but sweeter and better for eating raw and the males are better for cooking. I didn't know this!"", ""fb/david avocado wolfe Flip the bell peppers over to check their gender. The ones with four bumps are female and those with three bumps are male. The female peppers are full of seeds, but sweeter and better for eating raw and the males are better for cooking. I didn't know this!"", [('eng', 1.0)])]",['False information'],"}
         \item \texttt{fact check}: A collection of previously fact-checked claims from verified fact-checking organizations, for example :
 \textit{"(' Are avocados good for you?', ' Are avocados good for you?', [('eng', 1.0)])","[(1525653998.0, 'https://metafact.io/factchecks/175-are-avocados-good-for-you')]",}

 %
    
    


    \item \texttt{pair}: The mapping between fact check ids and post ids.
    \item \texttt{claim metadata}: Additional information about each fact-checked claim, including source, publication date, and veracity rating.  
    \item \texttt{relevant claims}: Human-identified relevant fact-checked claims that match the query, serving as gold standard matches.
\end{itemize}

The task is to retrieve relevant fact-checked claims from the \texttt{fact\_check} collection that match the given post. 
Retrieved claims are classified into the following two tracks based on language:

\begin{itemize}
    \item \texttt{monolingual}: Retrieving relevant fact-checked claims in the same language as the query.  
    \item \texttt{crosslingual}: Retrieving relevant fact-checked claims across different languages.
\end{itemize}

The provided training data consists of 153,743 samples in the \texttt{fact\_checks.csv}. The distribution across languages is:
English: 85,734 (55.76\%),
Portuguese: 21,569 (14.03\%),
Arabic: 14,201 (9.24\%),
Spanish: 14,082 (9.16\%),
Malay: 8,424 (5.48\%),
German: 4,996 (3.25\%),
French: 4,355 (2.83\%),
and Thai: 382 (0.25\%).
%

The test phase introduced two surprise languages: Turkish and Polish.

The agreement between the model's predictions and the ground truths is evaluated based on \textbf{success-at-10 (S@10)}, which measures if at least one relevant fact-checked claim appears among the top 10 retrieved results. Participants may choose to participate in either one or both tracks when making submissions, with a maximum of 5 submissions allowed during the test phase, where only the last submission will be counted for the final leaderboard. In our submission, we participated only in the \texttt{monolingual} track.

\section{Related Work}

Fact-checked claim retrieval has seen various approaches over the past few years. Early work by \cite{shaar-etal-2020-known} and later, the extensive MultiClaim study by ~\cite{Pikuliak_2023}, provided a broad multilingual dataset that expanded the task past English and monolingual settings. Their work highlighted the challenges in retrieving fact-checked claims across many languages.

Parallel to the fact-check retrieval, the development of Multilingual Language Models (MLLMs) is explained in \cite{doddapaneni2021primerpretrainedmultilinguallanguage}’s primer. MLLMs like mBERT and XLM-R have shown impressive zero-shot transfer capabilities in a wide range of tasks. These models are designed to learn shared representations across languages, using large-scale multilingual pre-training to enable cross-lingual transfer.


\section{System Overview}
The section presents a detailed view of our proposed system which generates the monolingual predictions for the given fact checks and posts; based on the given tasks file.

\subsection{Data Cleaning}
We decided to implement the data loading procedure given to us in the task, which utilized the \texttt{load.py} script. This script loads and preprocesses \texttt{posts.csv}, \texttt{fact\_checks.csv}, and \texttt{pairs.csv}. It checks if the files exist, and then handles newline characters in text fields by replacing "\texttt{\textbackslash n}" with escaped "\texttt{\textbackslash\textbackslash n}" before using \texttt{ast.literal\_eval} to parse string representations of lists or dictionaries. The script loaded \texttt{fact\_checks.csv} and \texttt{posts.csv} into Pandas DataFrames, filling missing values with empty strings and setting specific columns to be processed with the \texttt{parse\_col} function. The use of this script led us to having clean data with defined columns.

\subsection{Model Discussion}
The exploration of models was twofold; we first wanted to see the model performance on the dev set and then proceeded further to tune the model whose performance was found to be suitable.

We used the TF-IDF (Term Frequency-Inverse Document Frequency) model, implemented using scikit-learn’s \texttt{TfidfVectorizer}\footnote{ \url{https://scikit-learn.org/stable/modules/generated/sklearn.feature_extraction.text.TfidfVectorizer.html}.}. It is a statistical method used to convert text into numerical features by evaluating the importance of words in a document relative to a collection of documents. We used the TF-IDF model with modifications to the 'max\_features' (vector\_size) and 'analyzer' parameters. For the final system, only \texttt{task.json}, which is a task configuration file was used to ensure that the test data remained unseen.

We lay out a standard procedure and follow that for every model. Our procedure included the following steps:
    1) We start with parsing the tuple string format of the fact check data; and then proceed towards parsing the instances string to extract the URLs and timestamps. And to preprocess the post, we combine all the relevant text fields from a post.
    2) The \texttt{create\_retrieval\_system} function takes in the fact checks, task configuration file, and posts to create vector embeddings for the specific model. And then returns the retrieval model, fact check vectors, and fact check IDs for further retrieval tasks.
    3) The \texttt{retrieve\_fact\_checks} function takes an individual post's data, the model retriever (which calls the model), vector embeddings, fact check ids and the number of required results dictated by the organizers as input (10). The post's text data is converted into vector representations and similarities are computed between the query and the fact check. The fact checks are then ranked by similarity score and the list of top 10 fact check ids is returned by the function.
    4) The \texttt{generate\_predictions} function handles multiple posts at once. It processes each post by first extracting relevant text data, converting it into vector representations, and then using the retrieval model from \texttt{retrieve\_fact\_checks} to find the most similar fact checks. The function generates predictions by mapping each post to the corresponding top-ranked fact check IDs based on similarity scores. The output is returned in a JSON-like structure that maps multiple post IDs to all of their corresponding fact check IDs.
    5) In our pipeline the main function is responsible for loading the input data, processing each language individually, and generating predictions for each post. The processing includes parsing the posts' content, applying the retrieval system, and selecting relevant fact checks. The function then saves the predictions in a standard \texttt{monolingual\_predictions.json} file, which is used for evaluation on the task organizers' platform, CodaBench.

\begin{table*}
  \centering
  \caption{Performance scores across languages for different methods (Results on dev splits).}
  \label{tab:scores}
  \begin{tabular}{lccccccccc}
    \toprule
    \textbf{Sys.}             & \textbf{eng} & \textbf{spa} & \textbf{deu} & \textbf{por} & \textbf{fra} & \textbf{ara} & \textbf{msa} & \textbf{tha} & \textbf{avg} \\
    \midrule
    XLM-R                 & 0.0146       & 0.0488       & 0.0361       & 0.0497       & 0.0798       & 0.2436       & 0.0286       & 0.2857       & 0.0984      \\
    XLM-R2               & 0.0439       & 0.0667       & 0.0843       & 0.0762       & 0.1223       & 0.2692       & 0.0857       & 0.2857       & 0.1293      \\
    FastT                    & 0.1109       & 0.1805       & 0.1084       & 0.1788       & 0.2287       & 0.4359       & 0.2762       & 0.4286       & 0.2435      \\
    Distil                  & 0.2678       & 0.2585       & 0.1807       & 0.2417       & 0.4362       & 0.3846       & 0.3714       & 0.4286       & 0.3212      \\
        TFIDF-B              & 0.2678       & 0.4634       & 0.3855       & 0.4503       & 0.6330       & 0.7436       & 0.5619       & 0.9048       & \textbf{0.5513}      \\
    T5                     & 0.4812       & 0.5854       & 0.4337       & 0.5464       & 0.7128       & 0.7949       & 0.7238       & 0.8095       & 0.6360      \\
    E5                    & 0.4351       & 0.5821       & 0.4096       & 0.5331       & 0.7287       & 0.8205       & 0.7810       & 0.9048       & 0.6494      \\
    \bottomrule
  \end{tabular}
\end{table*}

\begin{table*}
  \centering
  \caption{Performance scores across languages for different TF-IDF configurations (Results on dev splits).}
  \label{tab:tfidf_scores}
  \begin{tabular}{lccccccccc}
    \toprule
    \textbf{Sys.}   & \textbf{eng} & \textbf{spa} & \textbf{deu} & \textbf{por} & \textbf{fra} & \textbf{ara} & \textbf{msa} & \textbf{tha} & \textbf{avg} \\
    \midrule
    Base          & 0.2678       & 0.4634       & 0.3855       & 0.4503       & 0.6330       & 0.7436       & 0.5619       & 0.9048       & 0.5513      \\
    C-WB           & 0.2908       & 0.3431       & 0.2410       & 0.4172       & 0.4681       & 0.6667       & 0.3619       & 0.8333       & 0.4528      \\
    Char              & 0.2971       & 0.3496       & 0.2651       & 0.4238       & 0.4787       & 0.6667       & 0.3714       & 0.8571       & 0.4637      \\
    10 K               & 0.5983       & 0.8065       & 0.6386       & 0.7947       & 0.8032       & 0.7821       & 0.8000       & 0.8810       & 0.7630      \\
    15 K               & 0.6130       & 0.8358       & 0.6627       & 0.8278       & 0.8032       & 0.7821       & 0.8000       & 0.8810       & \textbf{0.7757}      \\
    \bottomrule
  \end{tabular}
\end{table*}

\begin{table*}
  \centering
  \caption{Unseen Test Set Scores: Comparison between Duluth and the Top Ranked System}
  \label{tab:unseen_comparison}
  \begin{tabular}{lccccccccccc}
    \toprule
    \textbf{Sys.} & \textbf{pol} & \textbf{eng} & \textbf{msa} & \textbf{por} & \textbf{deu} & \textbf{ara} & \textbf{spa} & \textbf{fra} & \textbf{tha} & \textbf{tur} & \textbf{avg} \\
    \midrule
    DLH    & 0.626 & 0.452 & 0.8495 & 0.558 & 0.690 & 0.820 & 0.546 & 0.814 & 0.8415 & 0.686 & \textbf{0.6883} \\
    Top    & 0.926 & 0.916 & 1.000  & 0.926 & 0.958 & 0.986 & 0.974 & 0.972 & 0.9945 & 0.948 & 0.9601 \\
    \bottomrule
  \end{tabular}
\end{table*}


	
	
	
	
	

\section{Experiment}

We stuck to TF-IDF because it gave us the best results. However, there were several models that we wanted to experiment with before finalizing TF-IDF:

    \textbf{TF-IDF} –  Within TF-IDF, we experimented with different vector sizes and with the analyzer parameter having values: 'word', 'char', 'charwb'.
    
    \textbf{XLM-RoBERTa \cite{conneau-etal-2020-unsupervised}} – We tried two approaches for XLM-RoBERTa\footnote{ \url{https://huggingface.co/FacebookAI/xlm-roberta-base}.}, the first approach used mean pooling over token embeddings weighted by attention masks, capturing richer contextual information, whereas the second approach relied on the CLS token embedding. The first approach processed texts one at a time, making it slower, while the second implemented batch processing (batch size = 8), which aimed to improve efficiency.

    \textbf{Multilingual E5 Large \cite{wang2024multilingual}} – We use the E5 Retriever, a transformer-based embedding model\footnote{ \url{https://huggingface.co/intfloat/multilingual-e5-large}.}.
    Our retriever encodes text using an average pooling strategy over token embeddings, followed by L2 normalization for better similarity computation. We generated dense embeddings in batches.
    
    \textbf{FastText \cite{joulin2016fasttext}} – Fasttext uses pre-trained word vectors for 157 languages \cite{grave2018learning}  (\texttt{cc.en.300.bin})\footnote{ \url{https://fasttext.cc/docs/en/crawl-vectors.html}.}. Our FastTextRetriever preprocesses text by removing special characters and normalizing case before generating sentence embeddings. We compute dense representations for fact-check claims and social media posts, storing fact-check embeddings in batches.
    
    \textbf{T5-Model \cite{ni2021largedualencodersgeneralizable}} – The \texttt{sentence-transformers/gtr-t5-large}\footnote{ \url{https://huggingface.co/sentence-transformers/gtr-t5-large}.} was used to generate dense text embeddings. Our GTR Retriever preprocesses text by removing special characters and normalizing case before encoding claims and social media posts into 768-dimensional vectors.
    
    \textbf{distilBERT \cite{sanh2020distilbertdistilledversionbert}} – We utilize \texttt{distilbert-base-nli-stsb-mean-tokens}\footnote{\url{https://huggingface.co/hlyu/distilbert-base-nli-stsb-mean-tokens}.} to generate dense text embeddings. Our DistilBERTRetriever preprocesses text by removing special characters and normalizing case before encoding claims and social media posts into contextualized vector representations.

\subsection{Experimentation with TF-IDF}
When we finalized our model, we wanted to improve performance and thought the best way to do so would be to tinker with model parameters such as \texttt{max\_features} and \texttt{analyzer}.

The \texttt{max\_features} parameter controls the vector size by limiting the number of unique terms in the vocabulary. It retains only the topmost important words based on term frequency. Hence, the TF-IDF matrix will have at most that many columns, where each column represents a term.

The \texttt{analyzer} parameter controls how the input text is processed before applying the TF-IDF transformation. The default setting, \texttt{word}, tokenizes the text into individual words, allowing the model to capture word-level features. Other options, such as \texttt{char} and \texttt{char\_wb}, allowed for character-level tokenization or character n-grams.

The reason why we chose the default \texttt{word}, was to ensure that the most relevant features are captured at the word level, and because of all permutations and combinations of both model parameters, we were getting the best model performance and \textbf{success@10} by utilizing the default analyzer parameter.

\section{Experimental Results and Discussion}

The default parameter `word` along with a vector size of 15,000 gave us the best \textbf{success@K} score of 0.78 (avg) on the dev set and 0.69 (avg) on the test set.


The Duluth system achieved the highest success@K score of 0.78 (dev set) and 0.69 (test set) with the TF-IDF 15K configuration, which outperformed all other methods.

The TF-IDF 15K configuration showed consistent high performance across all languages, and improved over the TF-IDF 10K model (0.7757 vs. 0.7630 avg). TF-IDF performed well on English (0.6130) and Spanish (0.8358), which indicated that increasing the vocabulary size improved retrieval performance. However, character-based TF-IDF approaches underperformed, such as TF-IDF 15K (Char WB Analyzer) with an average score of 0.4528, which suggests that word-based tokenization was more effective than character-based representations.

 From the neural methods, E5 Large achieved an average score of 0.6494 and T5 Model closely followed at 0.6360. XLM-RoBERTa and DistilBERT performed poorly, with XLM-RoBERTa averaging only 0.0984 and DistilBERT at 0.3212.
 

TF-IDF was pretty standard across languages, but minor performance drops were observed for German (DEU) and Thai (THA).
Character-based analyzers underperformed, more so in German and Arabic, mainly due to their complex word forms and stuck together structures.

E5 Large and T5 Model had strong performance, but they had more errors in lower-resource languages (e.g., Malay and Thai), this could be due to pretraining biases as these models have largely been trained on other languages.

BERT-based models had very low performance, especially on Arabic (ARA), Malay (MSA), and Thai (THA), which suggested poor generalization in non-Latin scripts.


\section{Future Work and Conclusions}

Potential modifications to improve the performance of T5 Model and E5 Large could be fine-tuning these models on a more extensive and diverse set of fact-checking data to help them better capture domain-specific features.

Models could be generalized more by augmenting and balancing the training data, resulting in a reduced bias observed in pretraining.
Experimenting with language-specific tokenizers could address difficulty with non-Latin scripts and complex word forms.

We presented an optimized TF-IDF retrieval system for multilingual fact-checked claim retrieval. By fine-tuning vector dimensions and using word-level tokenization with a 15,000-feature vocabulary, our approach achieved robust performance—0.78 success@10 on the development set and 0.69 on the test set. These results highlight that, with careful tuning, traditional methods can be competitive even against advanced neural models, especially in high-resource settings.

We plan to investigate why our neural methods did not reach the level of TF-IDF by doing a case-by-case error analysis of where TF-IDF was performant and where it failed, to see if there exist any patterns to discover.

Future work could incorporate the reviewers' recommendation of a hybrid approach, combining TF-IDF and neural embeddings, to better support lower-resource languages and enhance retrieval effectiveness.

\section{Acknowledgments}
The authors would like to thank the organizers for
 the opportunity to participate in SemEval-2025
 Task 7 and the Department of Computer Science, University of Minnesota Duluth, for helping us with all resources needed to participate in this task. We also appreciate the valuable input
and guidance provided by the reviewers.

\bibliography{custom}

\begin{thebibliography}{11}
\providecommand{\natexlab}[1]{#1}

\bibitem[{Conneau et~al.(2020)Conneau, Khandelwal, Goyal, Chaudhary, Wenzek, Guzm{\'a}n, Grave, Ott, Zettlemoyer, and Stoyanov}]{conneau-etal-2020-unsupervised}
Alexis Conneau, Kartikay Khandelwal, Naman Goyal, Vishrav Chaudhary, Guillaume Wenzek, Francisco Guzm{\'a}n, Edouard Grave, Myle Ott, Luke Zettlemoyer, and Veselin Stoyanov. 2020.
\newblock \href {https://doi.org/10.18653/v1/2020.acl-main.747} {Unsupervised cross-lingual representation learning at scale}.
\newblock In \emph{Proceedings of the 58th Annual Meeting of the Association for Computational Linguistics}, pages 8440--8451, Online. Association for Computational Linguistics.

\bibitem[{Doddapaneni et~al.(2021)Doddapaneni, Ramesh, Khapra, Kunchukuttan, and Kumar}]{doddapaneni2021primerpretrainedmultilinguallanguage}
Sumanth Doddapaneni, Gowtham Ramesh, Mitesh~M. Khapra, Anoop Kunchukuttan, and Pratyush Kumar. 2021.
\newblock \href {https://arxiv.org/abs/2107.00676} {A primer on pretrained multilingual language models}.
\newblock arXiv preprint.
\newblock \emph{Preprint}, arXiv:2107.00676.

\bibitem[{Grave et~al.(2018)Grave, Bojanowski, Gupta, Joulin, and Mikolov}]{grave2018learning}
Edouard Grave, Piotr Bojanowski, Prakhar Gupta, Armand Joulin, and Tomas Mikolov. 2018.
\newblock Learning word vectors for 157 languages.
\newblock In \emph{Proceedings of the International Conference on Language Resources and Evaluation (LREC 2018)}.

\bibitem[{Joulin et~al.(2016)Joulin, Grave, Bojanowski, Douze, J{\'e}gou, and Mikolov}]{joulin2016fasttext}
Armand Joulin, Edouard Grave, Piotr Bojanowski, Matthijs Douze, H{\'e}rve J{\'e}gou, and Tomas Mikolov. 2016.
\newblock Fasttext.zip: Compressing text classification models.
\newblock \emph{arXiv preprint arXiv:1612.03651}.

\bibitem[{Ni et~al.(2021)Ni, Qu, Lu, Dai, Ábrego, Ma, Zhao, Luan, Hall, Chang, and Yang}]{ni2021largedualencodersgeneralizable}
Jianmo Ni, Chen Qu, Jing Lu, Zhuyun Dai, Gustavo~Hernández Ábrego, Ji~Ma, Vincent~Y. Zhao, Yi~Luan, Keith~B. Hall, Ming-Wei Chang, and Yinfei Yang. 2021.
\newblock \href {https://arxiv.org/abs/2112.07899} {Large dual encoders are generalizable retrievers}.
\newblock \emph{Preprint}, arXiv:2112.07899.

\bibitem[{Peng et~al.(2025)Peng, Moro, Gregor, Srba, Ostermann, Simko, Podroužek, Mesarčík, Kopčan, and Søgaard}]{semeval2025task7}
Qiwei Peng, Robert Moro, Michal Gregor, Ivan Srba, Simon Ostermann, Marian Simko, Juraj Podroužek, Matúš Mesarčík, Jaroslav Kopčan, and Anders Søgaard. 2025.
\newblock Semeval-2025 task 7: Multilingual and crosslingual fact-checked claim retrieval.
\newblock In \emph{Proceedings of the 19th International Workshop on Semantic Evaluation}, SemEval 2025, Vienna, Austria.

\bibitem[{Pikuliak et~al.(2023)Pikuliak, Srba, Moro, Hromadka, Smoleň, Melišek, Vykopal, Simko, Podroužek, and Bielikova}]{Pikuliak_2023}
Matúš Pikuliak, Ivan Srba, Robert Moro, Timo Hromadka, Timotej Smoleň, Martin Melišek, Ivan Vykopal, Jakub Simko, Juraj Podroužek, and Maria Bielikova. 2023.
\newblock \href {https://doi.org/10.18653/v1/2023.emnlp-main.1027} {Multilingual previously fact-checked claim retrieval}.
\newblock In \emph{Proceedings of the 2023 Conference on Empirical Methods in Natural Language Processing}, page 16477–16500. Association for Computational Linguistics.

\bibitem[{Sanh et~al.(2020)Sanh, Debut, Chaumond, and Wolf}]{sanh2020distilbertdistilledversionbert}
Victor Sanh, Lysandre Debut, Julien Chaumond, and Thomas Wolf. 2020.
\newblock \href {https://arxiv.org/abs/1910.01108} {Distilbert, a distilled version of bert: smaller, faster, cheaper and lighter}.
\newblock \emph{Preprint}, arXiv:1910.01108.

\bibitem[{Shaar et~al.(2020)Shaar, Babulkov, Da~San~Martino, and Nakov}]{shaar-etal-2020-known}
Shaden Shaar, Nikolay Babulkov, Giovanni Da~San~Martino, and Preslav Nakov. 2020.
\newblock \href {https://doi.org/10.18653/v1/2020.acl-main.332} {That is a known lie: Detecting previously fact-checked claims}.
\newblock In \emph{Proceedings of the 58th Annual Meeting of the Association for Computational Linguistics}, pages 3607--3618, Online. Association for Computational Linguistics.

\bibitem[{Sparck~Jones(1972)}]{sparck1972statistical}
Karen Sparck~Jones. 1972.
\newblock A statistical interpretation of term specificity and its application in retrieval.
\newblock \emph{Journal of documentation}, 28(1):11--21.

\bibitem[{Wang et~al.(2024)Wang, Yang, Huang, Yang, Majumder, and Wei}]{wang2024multilingual}
Liang Wang, Nan Yang, Xiaolong Huang, Linjun Yang, Rangan Majumder, and Furu Wei. 2024.
\newblock Multilingual e5 text embeddings: A technical report.
\newblock \emph{arXiv preprint arXiv:2402.05672}.

\end{thebibliography}




\end{document}